# *Taec*: a Manually annotated text dataset for trait and phenotype extraction and entity linking in wheat breeding literature

Claire Nédellec[1,*], Clara Sauvion[1], Robert Bossy[1], Mariya Borovikova[1,2], Louise Deléger[1]
[1] Université Paris-Saclay, INRAE, MaIAGE, Jouy-en-Josas, France
[2] TETIS, Univ. Montpellier, AgroParisTech, CIRAD, CNRS, INRAE, Montpellier, France
*Corresponding author
E-mail: claire.nedellec@inrae.fr

# Abstract

Wheat varieties show a large diversity of traits and phenotypes. Linking them to genetic variability is essential for shorter and more efficient wheat breeding programs. Newly desirable wheat variety traits include disease resistance to reduce pesticide use, adaptation to climate change, resistance to heat and drought stresses, or low gluten content of grains. Wheat breeding experiments are documented by a large body of scientific literature and observational data obtained in-field and under controlled conditions. The cross-referencing of complementary information from the literature and observational data is essential to the study of the genotype-phenotype relationship and to the improvement of wheat selection.

The scientific literature on genetic marker-assisted selection describes much information about the genotype-phenotype relationship. However, the variety of expressions used to refer to traits and phenotype values in scientific articles is a hinder to finding information and cross-referencing it. When trained adequately by annotated examples, recent text mining methods perform highly in named entity recognition and linking in the scientific domain. While several corpora contain annotations of human and animal phenotypes, currently, no corpus is available for training and evaluating named entity recognition and entity-linking methods in plant phenotype literature. The *Triticum aestivum trait Corpus* is a new gold standard for traits and phenotypes of wheat. It consists of 540 PubMed references fully annotated for trait, phenotype, and species named entities using the *Wheat Trait and Phenotype Ontology* and the species taxonomy of the *National Center for Biotechnology Information*. A study of the performance of tools trained on the *Triticum aestivum trait Corpus* shows that the corpus is suitable for the training and evaluation of named entity recognition and linking. This corpus is currently the most comprehensive resource for plant phenotype information.

# Introduction

The improvement of agricultural plant species has become an international concern due to the increasing demand for food to support a growing world population and the need to counteract the decline of resources, especially water and oil. This is especially true for wheat, the most widely grown crop after rice. Climate change and reduction in inputs (water, fertilizers, and pesticides) and acreages are new environmental constraints that lead to an increased need for wheat variety traits related to tolerance to water limitation and temperature stress, resistance to wind (e.g., stalk mechanical strength, resistance to lodging) and disease, or nutrient use efficiency (1). Wheat is used in a wide range of new food products, the demand for which is increasing due to changes in dietary habits and the search for health benefits. Wheat grain composition is therefore an important target for change: for example, low gluten preparations, textured proteins for vegetarian and low-carbohydrate products, starch composition involved in the preparation of baked goods, meat products and confectionery, and lower levels of synthetic chemical products in the design of breading, coating, and brine additive.

Therefore, a wide variety of wheat traits are being investigated, ranging from response to environmental conditions (biotic and abiotic), quality (for milling, for food), growth (e.g., yield, vigor, nutrient use), morphology, to reproduction.

The recent advent of genomic tools contributed to improving the linkage between molecular markers and genes of agronomic interest. This information is being integrated into increasingly shorter breeding programs to move from genetic toward genomic variety selection. Recently, many varieties and molecular markers have been developed for bread wheat (2). Data from breeding experiments are spread over thousands of heterogeneous datasets and publications (3). The goal of the D2KAB project (*Data to Knowledge in Agriculture and Biodiversity*; https://d2kab.mystrikingly.com/), of which this work is a part, is to develop new semantic web-based tools for the semantic description of agricultural data, making them actionable, and openly accessible, according to the FAIR (findable, accessible, interoperable, and reusable) principles (4). Breeders use the Crop Ontology to annotate phenotypic observation data (5).

However, the automatic annotation of textual data remains challenging because of the large number of traits and the broad diversity of the vocabulary used to designate them.

In addition, the scope of the traits differs depending on whether they come from texts or from field observations. Traits mentioned in documents usually pertain to the general properties of the wheat variety. On the other hand, observational data describe specific states of the plant within a limited spatial and temporal scope which must be aggregated and experimentally confirmed to derive the general characteristics of the observed wheat variety. The *Wheat Trait and Phenotype Ontology* (WTO; http://agroportal.lirmm.fr/ontologies/WHEATPHENOTYPE) was developed to meet the requirements for the annotation of textual data by ontologies (6), namely *named-entity linking* (NEL). WTO contains 596 trait and phenotype classes and covers all dimensions.

As shown in Table 1, the labels of the ontology classes and the textual expressions differ in many ways, which prevents the application of string matching for NEL. The same remark applies to the Crop Ontology. More powerful information extraction methods are needed to automate the recognition of the trait and phenotype entities in the text and link them to the corresponding classes in the WTO ontology.

**Table 1. Examples of ontology labels and text mentions.**

| Text entity | Class label |
| --- | --- |
| heading time | Ear emergence time |
| number of fertile florets at anthesis | percentage of florets without grain |
| highly resistant to leaf rust caused by *Puccinia triticina* | resistance to Leaf Rust |
| resistance to single isolates of *M. graminicola* | resistance to Septoria Leaf Blotch |
| HTAP resistance | high-temperature resistance |
| phenolic content | grain polyphenol content |
| number of tillers | shoot number per plant |
| TKW | thousand kernel weight |
| low molecular weight glutenin subunit | glutenin content |

Annotated corpora are needed to assess information extraction methods on their ability to recognize entities and their type, i.e., named-entity recognition (NER), and to predict the correct class, i.e., named-entity linking.

The current state-of-the-art algorithms for NER/NEL heavily rely on supervised Machine Learning methods. Although the number of training examples needed to train machine learning-based methods decreases with the advent of few-shot and zero-shot algorithms (7), the quality of prediction remains correlated with the availability of training examples of the target information.

NER and NEL corpora in Life Science mainly focus on human health. A few contain annotations of plant properties, but none of them provide annotations of traits with an ontology. We developed the *Triticum aestivum trait Corpus* (*TaeC*) to fill this gap. We selected PubMed (https://www.ncbi.nlm.nih.gov/pmc/) as the bibliographic source for scientific documents because it is fully open, the titles and abstracts of references are short and focused texts, and they contain a wealth of trait and phenotype mentions.

The entity types of *TaeC* include trait, phenotype, and species. Traits are observable characteristics such as the height of the plant or the resistance to a particular disease. The phenotypes are the values of the traits, *e.g.*, *1,2 m* as the *height value,* or *highly resistant to wheat blast* as the value for the *wheat blast resistance* trait. We used WTO as the reference resource for their annotation. Species are relevant entities here because many hybrid varieties obtained by crossing wild (e.g., *Aegilops tauschii,* an annual grass species) and cultivated plants (e.g., *Hordeum vulgare*, barley) are mentioned besides bread wheat (*Triticum aestivum*), durum wheat (*Triticum durum*), and their subspecies. We used the NCBI taxonomy (https://www.ncbi.nlm.nih.gov/taxonomy) as the reference resource for species standard annotation. The NCBI taxonomy is a worldwide recognized resource that is also relevant to the association of genes to phenotypes through its linked genetic sequence databases.

## Related work

*Taec* was designed based on an analysis of the existing corpora. In order to evaluate the suitability of the corpus and ensure that the tasks are feasible but not trivial, we selected a few named entity recognition and linking methods.

Most work on information extraction has focused on human and animal phenotypes and their linkage with genetic peculiarities and abnormalities. The popular *Phenotype-Gene Relations* corpus (PGR) (8) has been annotated with the *Human Phenotype Ontology* (HPO), which is a standard vocabulary for phenotypic abnormalities that occur in human diseases (9) using the machine learning-based NER tool IHP (*Identifying Human Phenotypes*) (10). HPO does not include regular traits such as eye color but only abnormal traits such as ocular albinism, which makes it irrelevant for plants. The Bacteria Biotope corpus (11) was annotated using the *OntoBiotope* ontology (12), a standard vocabulary for biotopes and phenotypes of microbes. Although *Ontobiotope* describes normal phenotypes, microbes are too different from plants for the corpus and ontology to be reusable even if some of the high-level classes are the same, e.g., biotic and abiotic stress response or morphology.

Plant biology has so far been relatively underrepresented as a topic for the BioNLP community. For *Arabidopsis thaliana*, there have been some recent Information Extraction initiatives, such as the *KnownLeaf* literature curation system (13) and the *SeeDev* reference corpus (14). The *SeeDev* corpus concentrates on seed development described at the molecular level, which is different from our subject. The *Knownleaf* corpus focuses on the regulatory mechanisms of leaf growth and development, and key genes related to relevant mutant phenotypes. Our goal differs from that of *Knownleaf* in that we do not consider wheat phenotypes as normal or abnormal with respect to a normal genetic reference, but we aim to account for the full diversity of trait values in wheat varieties. Moreover, plant traits in the *Knownleaf* corpus were described using terms from the *Phenotype, Attribute, and Trait Ontology* (PATO) (15) formally combined with plant parts and tissues from the *Plant Trait Ontology* (TO) (16), and the *Brenda Tissue Ontology* (17) according to the entity-attribute-value (EAV) model. For example, in the text, "The reduced leaf area in the hub1-1 mutant", the reduced leaf area phenotype is formalized as three distinct entities, area as the property, reduced as the value, and leaf as the plant part. For *TaeC*, we preferred textual annotations that encompass the plant part, the trait, and possibly its value rather than a formal distinction of the entities as the *Knownleaf* project does. Indeed, the goal of the *D2KAB* project is to facilitate the retrieval and reading of information for human users by using expressions that are as close as possible to their habits. The annotation of the D2KAB wheat data is done using WTO for text data and the Wheat Crop Ontology CO_321 for experimental data, which both meet this need. Indeed, in both ontologies, the trait classes gather the trait and the plant part it characterizes; for example, *grain color* is the label of WTO:0000141 and CO_321:0000037 trait classes.

Most of the corpora with species annotations are related to biomedicine, although some corpora include plant species LINNAEUS (18), COPIOUS (19), and S1000 (20) are the most preeminent among them. Most species of the LINNAEUS corpus reflect the PubMed source focus and are related to human health, model species, and microbes. COPIOUS documents from the Biodiversity Heritage Library include botanical species of the Philippine biodiversity. The S1000 corpus is dedicated to biodiversity surveys. Its botanical subset contains 125 scientific articles with 357 plant species annotations, among which only a dozen species are relevant to wheat breeding, e.g., *Zea mays*, *Triticum aestivum*, *Oryza sativa*.

# Materials and Methods

In this section, we describe how we constructed the *TaeC* gold standard. We present the annotation schema, our method for selecting documents, the annotation guidelines, and the annotation process. In order to assess the suitability of the corpus for the development of the NER and NEL methods, we selected the information extraction methods described in this section, which we then applied to the corpus.

## Annotation Schema

The annotation scheme consists of three entity types, Species, Trait and Phenotype. Species are normalized by the NCBI taxonomy. Trait entities are normalized with the trait subtree (WTO_0000006) classes of WTO ontology. Phenotype entities are normalized with the *phenotype* subtree (WTO_0000005) classes of WTO ontology. Fig 1 shows an example where *earliness* is an entity of trait type, and *wheat* is an entity of species type.

**Fig. 1 Example of annotation in Taec corpus.**

The class associated with *earliness* is *plant precocity* (WTO_0000100). The class associated with *wheat* is *Triticum aestivum* (TaxID_4565).

## Document Selection

The documents of TaeC were selected using the PubMed query in Table 2.

**Table 2. Document selection query to PubMed.**

| |
|---|
| (("biomarkers"[MeSH Terms] OR "biomarkers"[All Fields] OR "marker"[All Fields] OR "markers"[All Fields])  AND ("genes"[MeSH Terms] OR "genes"[All Fields] OR "gene"[All Fields])  AND ("triticum"[MeSH Terms] OR "triticum"[All Fields] OR "wheat"[All Fields] OR "wheat s"[All Fields] OR "wheats"[All Fields]))  AND ((fha[Filter]) AND (english[Filter])) |

The first part of the query, with the marker and gene keywords (3 first lines), selects documents on wheat breeding and excludes documents on other uses of wheat, such as food processing or composition. Lines 4 and 5 of the query select documents about wheat species. The last query part selects documents in English and containing an abstract.
The query carried out on April 22, 2022, retrieved 5,596 documents, mostly published after 2011. To select a representative subset to be annotated, we applied the *AlvisNLP* workflow dedicated to bread wheat NER and NEL (21) to annotate the trait entities in titles and abstracts and link them to WTO. We then generated subsets of 6 documents where the selection is biased so that each set contains at least one document with entities annotated by one of the six main subtrees of WTO: development, growth, morphology, quality, reproduction, and response to environmental conditions. A document may be annotated by more than one subtree.

## Annotation guidelines

We prepared annotation guidelines by adapting the annotation guidelines of (21): we removed sections about irrelevant relationships and entities, and added new sections for entity linking of traits, phenotypes, and species. It resulted in an eight-page document that defines the entities, gives examples and counter-examples of annotations, and details exceptions and borderline cases (22).

## Annotation process and tools

The corpus entities pre-annotated by the *AlvisNLP* workflow dedicated to bread wheat were provided to the experts to save time. Two annotators successively annotated 540 documents; the first annotator had previous experience in manually annotating biomedical documents and performed an extensive annotation, the second annotator was an expert in biology and thoroughly reviewed all annotations. Issues were addressed in collaboration with two textual annotation experts and two external experts in wheat agronomic traits and disease resistance, which were the most complex traits to annotate.

For manual text annotation, we used the *AlvisAE* editor (23). It supports ontological annotation. It is implemented as a Web application, thus facilitating the participation and collaboration of domain experts. Fig 2 shows a screenshot of the main window of the *AlvisAE* wheat instance.

**Fig. 2. The *AlvisAE* annotation editor used to annotate *TaeC* manually.**

Automatic consistency checking was regularly performed to ensure high-quality annotation. The rules to be checked were (1) all entity mentions are tagged by classes, (2) the same classes tag identical entity mentions, and (3) there is no entity with both annotated and unannotated occurrences. The rules may not be relevant in some cases, i.e., a given expression may have different meanings depending on the context. Then consistency checking was therefore not used as an automatic revision tool but was intended to warn the annotators.

## Information extraction methods

The suitability of manually annotated corpora is estimated by applying baseline information extraction methods. We considered rule-based and machine-learning methods for named entity recognition and linking in *Taec*.

**Rule-based methods**
Previous work on the recognition and linking of wheat phenotype entities in texts includes the application of the *ToMap* method (24) using WTO (6). Candidate entities were first extracted by the *YaTeA* term extractor (25). The ToMap method then computes the similarity between candidate entities and candidate class labels based on their syntactic structure and common words in the same spirit as *MetaMap* (26). The information on the syntactic structure is particularly useful for processing complex prepositional phrases, which are common in the wheat breeding literature; for example, the complex phrase *the percentage of fertile florets at anthesis* should be mapped to the class WTO: 0000173 *percentage of florets without grain*. ToMap computes the correspondence between their main chunks that are the same, [percentage [of [florets]]], while their surface forms differ significantly. The preliminary experimental results published about the application of ToMap to the recognition and linking of wheat phenotype entities in texts were encouraging (21). We used it therefore for the joint NER and NEL tasks on *Taec*.

*Machine-learning-based methods*
Recent successful NER and NEL machine-learning methods are based on neural networks and language models. The recognition and linking tasks are both treated as classification tasks. Common NER approaches start from a general language model that is fine-tuned to the particular task using training examples. One prominent example is the RoBERTa model (27), which has become widely used due to its ability to be fine-tuned for multiple tasks, including NER (28). In the biological domain, the BioBERT model (29) is used for both NER (30); (31) and NEL (32) tasks with an additional fine-tuning stage. In our experiments, we employ the classic fine-tuning approach applied to RoBERTa and BioBERT models, with the addition of a dense classification layer and a softmax function, as introduced in (33) for classification with BERT language model.

Machine learning approaches combined with semantic ontology representations are prevalent for NEL on biomedical data. The C-Norm algorithm (34) achieves state-of-the-art performances on the Bacteria Biotope dataset. The method represents terms in the texts using

Word2vec embeddings (35) and ontology concepts using vectors that integrate hierarchical information from the ontology (36). It combines a single-layer feedforward neural network and a shallow convolutional neural network.

Another notable NEL algorithm in the biomedical domain is BioSyn (37). The authors of this method train a dense entity representation space using the BioBERT model. It uses synonym marginalization techniques as the objective function for training and a similarity function, leveraging top-k similar candidates to iteratively update model parameters.

For the NEL task involving traits and phenotypes, we use the C-Norm and BioSyn algorithms. Finally, we explore an end-to-end solution by applying the best NEL model to the predictions of the best NER model.

**Species recognition and linking**

Species recognition has inspired a significant amount of work. Most of the methods are dictionary-based, enhanced by additional local variation rules (e.g., Levenstein distance), contextual rules or intrinsic document word disambiguation, although there is growing interest in machine-learning-based solutions; see (38) for a survey. The limited range of species in the corpus, i.e., 106, does not require such advanced tools for their recognition and linking as in the general open domain.

As the NEL task for *Taec* is to link the entities to the NCBI taxonomy, we used this taxonomy as a dictionary for the prediction and linking of the entities. We selected the *AlvisTaxa* tool that achieved good performance in microbe species NER and NEL (39). It uses rewriting rules to generate species name variations, including abbreviations and acronyms, and performs contextual disambiguation.

# Results

This section presents the details of *TaeC,* and the NER/NEL results that we obtained from our experiments with state-of-the-art information extraction methods. Table 3 gives entity annotation and class frequencies of the corpus per type.

**Table 3. Figures of the *TaeC* corpus.**

| | | | |
|---|---|---|---|
| # Manually annotated documents | 540 | | |
| # Token | 142 726 | 264 per document on av. | |
| # Occurrences of phenotype | 1 598 | 2,96 per document on av. | 977 unique |
| # Occurrences of trait | 5 453 | 10.1 per document on av. | 1 696 unique |
| # Classes of traits or phenotypes | 233 | 11 forms per class on av. | |
| # Occurrences of plant species | 3 584 | | |
| # Taxa | 3 697 | 6,8 per document on av. | 278 unique |
| # Classes of taxon | 106 | 2,6 forms per class on av. | |
| # Total occurrences | 10 635 | | |

The high density of 13 trait and phenotype annotations per document confirms the relevant choice of PubMed references. The diversity of the mentions is shown by the high number of unique expressions compared to the number of occurrences. Each phenotype mention is repeated 1,6 times on average (1598/977), and each trait mention is repeated 3,2 times on average (5453/1696).

Trait and phenotypes entities are tagged by 233 different classes, yielding 11,4 unique mentions and 30,3 occurrences per class on average, which is high. As might be expected, the distribution varies significantly by class; see examples (1) and (2).

(1) the WTO:0000072 *class culm length* has four forms in the corpus, *culm length*, *stem length,* and their CL and SL acronyms,
(2) while WTO:0000146 *grain protein content* presents 45 different forms.

The diversity of taxon names is lower, with 2,6 forms per class on average. The number of occurrences (6,8 per document) is surprisingly high, but the occurrences are highly repeated (13,3 times on average), especially the word *wheat* with 1595 occurrences.

**Trait and phenotype annotation**

The trait and phenotype entities are adjectival or nominal expressions. Noun phrases are the most frequent, but adjectives also frequently express phenotypes (e.g., *sprouting-resistant*, *salt-tolerant*).

Many of them are acronyms (e.g., *GPC* standing for *Grain Protein Content*); prepositional phrases are common (e.g., *ratio of the quantity of glutenin to those of gliadin*, *number of grains per spike*); some forms include parenthesized expressions (e.g., *number of flowering branches (spikelets) per node*) where the term into parenthesis can be an acronym (e.g., *Efficient phosphate (Pi) uptake*, *responses to low red light/far-red light (R/FR) ratios*). Discontinuous entities are rare. For instance, in the text resistant to both WSMV and Triticum mosaic virus, two discontinuous annotations of phenotypes are made, i.e., *resistant to WSMV*, and *resistant to* Triticum *mosaic virus*.

The class labels differ from the entities in many ways, requiring deep domain expertise. First, the plant parts in trait and phenotype names can be designated by different names (e.g., *grain*/*kernel*/*seed*; *spike*/*ear; culm*/*stem*). WTO ontology defines alternative names for those traits but not all. The resistance to diseases (e.g., *resistance to wheat blast*) can be expressed as the resistance to the pathogen agent (e.g., *resistance to Magnaporthe grisea*). Fungi are the main cause of diseases in wheat. Beyond their official names, alternative names for separate morphs, teleomorph, anamorph, and holomorph are used. WTO ontology records many of them. For instance, eyes spot disease is caused by *Helgardia herpotrichoides* (syn *Pseudocercosporella herpotrichoides, Ramulispora herpotrichoides, Tapesia yallundae,* Cercosporella herpotrichoides). The corresponding resistance trait class *resistance to eyes spot* has therefore five alternative names accordingly.

The number of alternative names increased with the number of pathogen agents for a given disease, and the knowledge about causal agents evolves over time. Experts may disagree on the causal agents of diseases and species distinction. For instance, the synonyms of WTO:0000510 *resistance to wheat blast* include *resistance to Magnaporthe grisea*, *resistance to Magnaporthe oryzae*, and *resistance to Pyricularia grisea* which are considered different species by some experts, notably NCBI. The contribution to *Taec* annotation of experts in the fungi diseases of wheat has been significant in handling this issue.

Long noun phrases with subordinate clauses, infinitives, participles, *gerunds*, and verbless clauses are hopefully rare but more frequent than in other corpora. (1), (2), and (3) are illustrative examples.
(1) *Resistance to the disease septoria tritici blotch caused by the fungus Mycosphaerella graminicola*
(2) *resistance against a number of other important P. graminis f. sp. tritici pathotypes*
(3) *impact on frost damage in cereal reproductive tissues by influencing accumulation of genuine tolerance*

The *Taec* corpus includes paraphrases that reword traits and phenotypes in longer forms to make the meaning clearer and more precise. Examples of paraphrase are (4) and (5).
(4) *level of physical attachment of glumes to the rachilla of a spikelet* stands for *glume tenacity*.
(5) *grain yields of evaluated cultivars growing in the field under water-limited conditions*, stands for *drought tolerance*.

The treatment of such linguistic phenomena may be beyond the capabilities of usual entity recognition and linking tools.

The distinction between the trait and the method used to measure its value is another source of ambiguity. For example, grain weight traits are usually qualified by the method, e.g., *thousand *grain weight*, *test weight*. In this case, we include the name of the method in the annotation span. More complex cases are expressions such as *Zeleny sedimentation value*, mostly used in place of the trait name, *flour sedimentation volume* (related to *grain protein content*). The decision was also to annotate them as trait entities.

The annotation of recent articles in the corpus revealed the evolution of the domain due to the phenotyping method advances. New traits are studied, and new terms for denoting them occur. Examples of such new topics are related to plant architecture (tillers and root), physiology (hormone response, source-sink relationship), and reproduction. We extended WTO with 67 new specific classes to reflect the domain evolution. The new version is available at http://agroportal.lirmm.fr/ontologies/WHEATPHENOTYPE.

Consistency checking reveals manual annotation errors, among which the most frequent ones were over-specific or over-general class annotations of infrequent expressions because the size of WTO is too large for the experts to remember them all. These errors were automatically detected and manually corrected.

**Taxon annotation**

Species are equally designated by their scientific name (*e.g.*, *Triticum aestivum* L.), sometimes abbreviated (*e.g.*, *T. aestivum*), and by their vernacular name (*e.g.,* bread wheat, barley, rice). Vernacular names can be ambiguous, especially *bread wheat* which is frequently wrongly called *wheat*, although *wheat* is the vernacular general name of the *Triticum* genus that includes 19 species, not only *Triticum aestivum* species.

***Taec* format and distribution**

As usual for annotated corpora, *Taec* distribution includes (1) the text itself and its bibliographic metadata, (2) the entities, their position, and their type, and (3) the identifier of the ontology class.

It is provided in the BioNLP-ST standoff annotation format (40), which is suitable for representing entities linked to a semantic reference (Table 4). The text, the entities, and the semantic references are in separate files. The files are linked through the name of the file that plays the role of identifier.

**Table 4. File.txt contains the text of the document. File.a1 contains the named entity annotations, their identifier, their type, and their position. File.a2 contains the semantic annotations, the name of the reference, and the reference identifier.**

---

**File.txt**
*Efficiently tracking selection in a multiparental population: the case of earliness in wheat.*

**File.a1**
T1  Trait      75    83    earliness
T2  Species    88    92    wheat

**File.a2**
N1    NCBI_Taxonomy Annotation:T1 Referent: 4565
N3    WTO Annotation:T2 Referent:WTO: 0000100

---

The annotation set is split into three subsets, a train, a development, and a test subset with similar entity distributions. The train and development subsets are made public. The test set will be used in a future shared task and remains hidden to prevent overtraining. *TaeC* is available under CC-BY-ND License at:
https://entrepot.recherche.data.gouv.fr/dataset.xhtml?persistentId=doi:10.57745/GCYG3Q.

## Named entity recognition and linking

We evaluated the performance of state-of-the-art rule-based and machine-learning methods on the *Taec* corpus with the goal of estimating the difficulty of the NER and NEL tasks. We used the same metrics as the BioNLP Shared Task (40). In addition to the usual recall and precision metrics, we used strict matching and relaxed matching to compute entity recognition scores. Relaxed matching is measured by a variant of the Jaccard index applied to segments: the similarity of entity pairs is measured as the ratio between the size of the overlapping text segments and the size of the two merged text segments.

To avoid counting substitution errors twice, we used the Slot Error Rate (SER) that has been devised to undertake this shortcoming (41). For entity linking evaluation, we used strict matching and Wang's similarity (42) to compute hard and soft semantic annotation scores. Wang's similarity takes into account the number of common ancestors between the predicted label and the label to be predicted with closer terms contributing more to the similarity than more distant terms.

## Rule-based method

The recognition of phenotypes and traits and their linking to WTO was achieved by the *ToMap* rule-based method (24) described in the Related Work section above, which we find suitable for accurately identifying the borders of complex syntactic structures without training.

The species were recognized and linked to the NCBI taxonomy by the taxon recognition component *AlvisTaxa* based on dictionary-based matching (39). We found the method appropriate given that species names occurring in *TaeC* are both scientific and vernacular names and include abbreviations, and the method achieved good performances in Life Science tasks (11). Both components are part of the *AlvisNLP* workflow dedicated to bread wheat that we used for the pre-annotation. The results are reported in Table 5.

**Table 5. NER+NEL performances of the rule-based methods *AlvisTaxa* and *ToMap* on *TaeC*.**

|  | Strict match | | | | Jaccard index and Wang distance | | | |
|---|---|---|---|---|---|---|---|---|
|  | Precision | Recall | F1 | SER | Precision | Recall | F1 | SER |
| **Species (AlvisTaxa)** | 0,91 | 0,80 | 0,85 | 0,28 | 0,96 | 0,84 | 0,90 | 0,18 |
| **Trait (ToMap)** | 0,44 | 0,27 | 0,33 | 1,07 | 0,57 | 0,34 | 0,43 | 0,81 |
| **Phenotype (ToMap)** | 0,07 | 0,04 | 0,05 | 1,53 | 0,16 | 0,10 | 0,12 | 1,33 |
| **All** | 0,59 | 0,41 | 0,49 | 0,87 | 0,68 | 0,48 | 0,56 | 0,67 |
| **Phenotypes=Traits** | 0,37 | 0,22 | 0,28 | 1,16 | 0,55 | 0,33 | 0,41 | 0,80 |

The prediction of *AlvisTaxa* is high with 0,85 F-measure. The analysis of false negatives shows that non-standard abbreviations such as *Th. Elongatum* were not recognized, either partly recognized, e.g., *Triticum aestivum* instead of Triticum aestivum L, or assigned to the wrong NCBI class, e.g., *Triticum turgidum* L. **var** durum known as *Triticum turgidum* **subsp**. durum. A preprocessing problem makes the names into parentheses unrecognizable (e.g., (*T. aestivum*)).

The poor prediction of correct boundaries of traits and phenotypes partly explains the medium performances, for example, the prediction of *spikelet number* instead of *total spikelet number* *per spike*. The performance increases with the relaxation of the boundaries as measured by the Jaccard index (right part of the table). The very low performances of phenotype prediction may also be due to the boundary issue. Many wrongly predicted phenotypes were predicted as traits with a substring of the reference string. An example is the prediction of the expression *grain quality* as a trait instead of predicting *decreased* **grain quality** as a phenotype. A similar example is the prediction of the *protein content* trait instead of the *lower* **protein content** phenotype. We measured the confusion impact in the bottom line of the table that shows the performance when traits and phenotypes are mixed.

**Machine learning methods**

The methods were all trained using the train and development subsets. We evaluated their performances by 10-fold cross-validation measured by micro-average of precision, recall, and F1.

**Named entity recognition**
For the NER task evaluation, we fine-tuned RoBERTa (27) and BioBERT (29) models. Table 6 presents their precision, recall, and F1 performances, as well as the Jaccard index measure.

Table 6. Performance evaluation of RoBERTa and BioBERTa methods for the recognition of phenotype, trait, and species entities.

| Method | Entity | Precision | Recall | F1 | Jaccard index |
|---|---|---|---|---|---|
| RoBERTa | Overall | **0.67** | **0.33** | **0.44** | **0.50** |
|  | Phenotype | **0.92** | **0.09** | **0.17** | **0.26** |
|  | Trait | 0.58 | **0.31** | **0.41** | **0.47** |
|  | Species | **0.79** | **0.44** | **0.57** | **0.62** |
| BioBERT | Overall | 0.61 | 0.29 | 0.39 | 0.46 |
|  | Phenotype | 0.84 | 0.03 | 0.05 | 0.15 |
|  | Trait | **0.6** | 0.29 | 0.39 | 0.44 |
|  | Species | 0.62 | 0.39 | 0.48 | 0.54 |

Both methods perform well considering the difficulty of the phenotype and trait recognition task. The recognition of species is surprisingly high, given that the algorithms do not use a species dictionary. RoBERTa achieved significantly better performances than BioBERT (in bold).

The confusion between phenotype and trait types could explain both methods very low performance for phenotype recognition. To test this hypothesis, we measure the performances of the models without distinguishing between both types. We call *Characteristics* the union of the two types. The experimental results are shown in Table 7.

Table 7. Performance evaluation of RoBERTa and BioBERTa methods for the recognition of characteristics and species entities.

| Method | Entity | Precision | Recall | F1 | Jaccard index |
|---|---|---|---|---|---|
| RoBERTa | Overall | 0.71 | 0.35 | 0.47 | 0.51 |
| | Characteristics | 0.67 | 0.31 | 0.42 | 0.49 |
| | Species | 0.77 | 0.44 | 0.56 | 0.62 |
| BioBERT | Overall | 0.63 | 0.32 | 0.43 | 0.47 |
| | Characteristics | 0.61 | 0.28 | 0.38 | 0.46 |
| | Species | 0.66 | 0.4 | 0.50 | 0.58 |

The experimental results confirm our hypothesis. When traits and phenotypes are combined into a single category, the algorithms demonstrate improved performance compared to when treated as separate categories. The overall performance of both methods significantly increases by + 3 points for RoBERTa and +4 points for BioBERT. It is worth noticing that not only traits and phenotypes are better recognized, but also species. These preliminary results encourage to the evaluation of the methods for the merged type in the future.

**Named entity linking**

Regarding the NEL evaluation, we assessed the performance of C-Norm (34) and BioSyn (37) for the prediction of trait and phenotype classes. We did not evaluate the prediction of species because the size of the NCBI taxonomy is beyond the capacity of the algorithms. The two models were provided with the gold-named entity for training and inference. The results are reported in Table 8.

Table 8. Performance evaluation of BioSyn and C-Norm NEL methods on phenotype and trait class prediction.

| Method | Entity | Precision | Recall | F1 | Wang similarity |
|---|---|---|---|---|---|
| C-Norm | Overall | 0.83 | 0.81 | **0.82** | **0.91** |
| | Phenotype | 0.88 | 0.82 | 0.85 | 0.87 |
| | Trait | 0.84 | 0.82 | **0.83** | **0.92** |
| BioSyn | Overall | 0.84 | 0.82 | 0.83 | 0.90 |
| | Phenotype | 0.92 | 0.87 | **0.89** | **0.89** |
| | Trait | 0.84 | 0.81 | 0.82 | 0.9 |

Both methods perform similarly on trait and phenotype types and achieved good scores compared to similar biomedical normalization tasks such as Bacteria Biotope'19 BB-Norm (11).

**Named entity recognition and linking**

The previous experiments to infer the classes to be linked were done using the gold reference entities. In this section, we present the results obtained by successively using the best method RoBERTa for NER and applying the best method C-Norm to RoBERTa results for NEL. The performances obtained are displayed in Table 9.

Table 9. Performance of combined NER and NEL methods RoBERTa + C-Norm on phenotype and trait entities.

| Entity | F1-micro | Wang similarity |
|---|---|---|
| Phenotype | 0,03 | 0,21 |
| Trait | 0,50 | 0,78 |
| Overall | 0,49 | 0,77 |
| Overall (Phenotype ∪ Trait) | 0,55 | 0,82 |

As expected, the final scores decreased compared to NEL applied to gold entities. The bad recognition of the phenotype entities has a great impact with an F1 score equal to 0,03; however, the Wang similarity at 0,21 for phenotypes shows that the predicted classes are not exact, but a significant part of them belong to the ancestors of the class to predict. The same comment can be made more generally by comparing the F1 scores and the Wang similarity measures that differ by almost 30 points. These encouraging results lead us to conclude that there is considerable room for improvement in using more effective and better-adapted methods.

*Methods and parameters*

The AlvisNLP bread wheat workflow is available at : https://forgemia.inra.fr/migale/wheat-tm. It includes the wheat-specific lexica of ToMap.

The code of the ToMap method is available at https://github.com/Bibliome/alvisnlp/tree/master/alvisnlp-bibliome/src/main/java/fr/inra/maiage/bibliome/alvisnlp/bibliomefactory/modules/tomap

C-Norm and BioSyn were run with the default parameters provided by the authors.

The values of the hyperparameters of RoBERTa, BioBERT, BioSyn and C-Norm were as follows:

- #epoch: 10
- Learning rate: 5e5
- Gradient accumulation step: 2
- Random seed: -42
- Maximum gradient norm: 10

# Discussion

The entity types of *TaeC* do not include varieties or cultivars despite their high relevance. This is left for future work because of its complexity. Wheat varieties may be referred to by their commercial names (e.g., *European Plant variety database*; https://ec.europa.eu/food/plant-variety-portal/), making them easy to recognize but not of great value to *TaeC*. More frequent and valuable for designing a gold corpus are mentions of cultivars or accessions obtained by genetic modification or crossing. Unfortunately, they are often not named entities, but more complex expressions, including verbal ones. Table 10 shows four illustrative examples of such terms. Their manual annotation requires a high level of expertise and resources at an unaffordable cost.

**Table 10. Examples of cultivar names and descriptions.**

[..] F2 and F2:3 lines derived from the cross of the multi-spikelet female 10-A and the uni-spikelet male BE89 [PMID: 32549690]

> [..]a F(6:7) recombinant inbred line (RIL) population by crossing Wangshuibai with the scab-susceptible cultivar Nanda2419 [PMID: 15290053]
>
> [..] wheat cultivars with Pina-null/Pinb-null allele [PMID: 24011219]
>
> The wheat accession H9020-1-6-8-3 is a translocation line previously developed from interspecific hybridization between wheat genotype 7182 and *Psathyrostachys huashanica*. [PMID: 30727301]

# Conclusions

In this paper, we have presented the rationale for the construction of the new reference corpus *Taec* of 540 documents relevant to wheat breeding research, and more generally to cultivated plant phenotyping. *Taec* is annotated with three entity types, i.e., traits, phenotypes, and species, and each entity is linked to a reference class from NCBI taxonomy for species and WTO for traits and phenotypes, yielding 10,635 occurrences of 340 classes. Our experimental results show that *Taec* is useful for training and evaluating NER and NEL methods, the scores are similar to those obtained on phenotypes of microbes and humans.

In long-term future work, we will consider variety, gene, and marker annotation to address the need to bridge the phenotypic and genetic data for integrated marker-assisted plant selection.

In short-term work, the NER/NEL methods we trained with Taec will be integrated into an AlvisNLP workflow that will be used as a service of the D2KAB project to automatically annotate all PubMed references on wheat. The wheat knowledge base under development in the *D2KAB* project will combine these textual annotations and the annotations of the Wheat Information System indexed by the *Crop Ontology CO_321* (http://www.wheatis.org/). The alignment of *WTO* and *CO_321* ontology classes will allow the user to query the observation data and the documents in a seamless way and take full advantage of the complementarity of the two sources.

# Acknowledgments

The authors thank Léonard Zweigenbaum (INRAE) for his contribution to the annotation and, Thierry Marcel and Jacques Le Gouis of INRAE for their contribution to the technical questions relating to the description of the phenotyping of wheat.
The authors thank the Migale platform for providing the resources to run AlvisNLP services (MIGALE, INRAE, 2020.
Migale Bioinformatics Facility, doi: 10.15454/1.5572390655343293E12).